\documentclass[letterpaper, 10 pt, conference]{IEEEtran}  

\IEEEoverridecommandlockouts                              

\usepackage{graphics} 
\usepackage{graphicx}
\usepackage{subfigure}
\usepackage{amsmath} 
\usepackage{amssymb}  
\usepackage[colorlinks=true, linkcolor=black, citecolor=black, urlcolor=black]{hyperref}
\usepackage{cleveref}
\usepackage[linesnumbered,ruled,vlined]{algorithm2e}
\usepackage{cite}

\title{\LARGE \bf
FACT: Fast and Active Coordinate Initialization for Vision-based Drone Swarms
}

\author{Yuan Li$^{1,2,\dag}$, Anke Zhao$^{1,2,\dag}$, Yingjian Wang$^{1,2,\dag}$, Ziyi Xu$^{1,2}$, Xin Zhou$^{1,2,*}$,\\ Jinni Zhou$^3$, Chao Xu$^{1,2}$, and Fei Gao$^{1,2,*}$
\thanks{$^\dag$Equal contributors.}
\thanks{$^1$Institute of Cyber-Systems and Control, College of Control Science and Engineering, Zhejiang University, Hangzhou 310027, China.}%
\thanks{$^2$Huzhou Institute of Zhejiang University, Huzhou 313000, China.}
\thanks{$^3$The Hong Kong University of Science and Technology (GZ).}
\thanks{$^*$Corresponding authors: Xin Zhou and Fei Gao.}%
\thanks{E-mail:\{yuanli\_cse, iszhouxin, fgaoaa\}@zju.edu.cn}%
}

\begin{document}
\maketitle
\thispagestyle{empty}
\pagestyle{empty}

\begin{abstract}
Swarm robots have sparked remarkable developments across a range of fields. While it is necessary for various applications in swarm robots, a fast and robust coordinate initialization in vision-based drone swarms remains elusive. To this end, our paper proposes a complete system to recover a swarm's initial relative pose on platforms with size, weight, and power (SWaP) constraints. To overcome limited coverage of field-of-view (FoV), the drones rotate in place to obtain observations. To tackle the anonymous measurements, we formulate a non-convex rotation estimation problem and transform it into a semi-definite programming (SDP) problem, which can steadily obtain global optimal values. Then we utilize the Hungarian algorithm to recover relative translation and correspondences between observations and drone identities. To safely acquire complete observations, we actively search for positions and generate feasible trajectories to avoid collisions. To validate the practicability of our system, we conduct experiments on a vision-based drone swarm with only stereo cameras and inertial measurement units (IMUs) as sensors. The results demonstrate that the system can robustly get accurate relative poses in real time with limited onboard computation resources. The source code is released. 
\end{abstract}

\section{INTRODUCTION}
\label{sec1}
Swarm robots have proven transformative across domains such as exploration~\cite{MeetingMergingMission}, search~\cite{searching}, and surveillance~\cite{surveillance}, embodying unmatched versatility. To ensure the successful execution of these collaborative tasks, it is essential for each robot within swarms to accurately estimate the relative poses to other robots. The lack of external positioning equipment in environments such as caves or indoor areas restricts the input sources to onboard sensors only. This makes relative positioning a challenging task for robots with size, weight, and power (SWaP) constraints, such as micro drones, as they are limited in the number and weight of sensors. Given these hurdles, it is essential to enable self-localization and relative pose recovery based solely on lightweight onboard sensors. 

For SWaP-constrained robots, a stereo camera along with an inertial measurement unit (IMUs) is shown to be the minimum sensor set required for high-performance aerial swarm navigation~\cite{zhou2022swarm, ego-swarm}. The stereo-vision-based setting is one of the most successfully validated sensor settings, especially on SWaP-constrained robots~\cite{depthcamera2021fish, depthcameraarnold2018search}. It enables efficient capture of three-dimensional coordinate information upon detecting other robots, providing a comprehensive set of bearing and distance information.

\begin{figure}[t]
  \centering
  \includegraphics[width=1\linewidth]{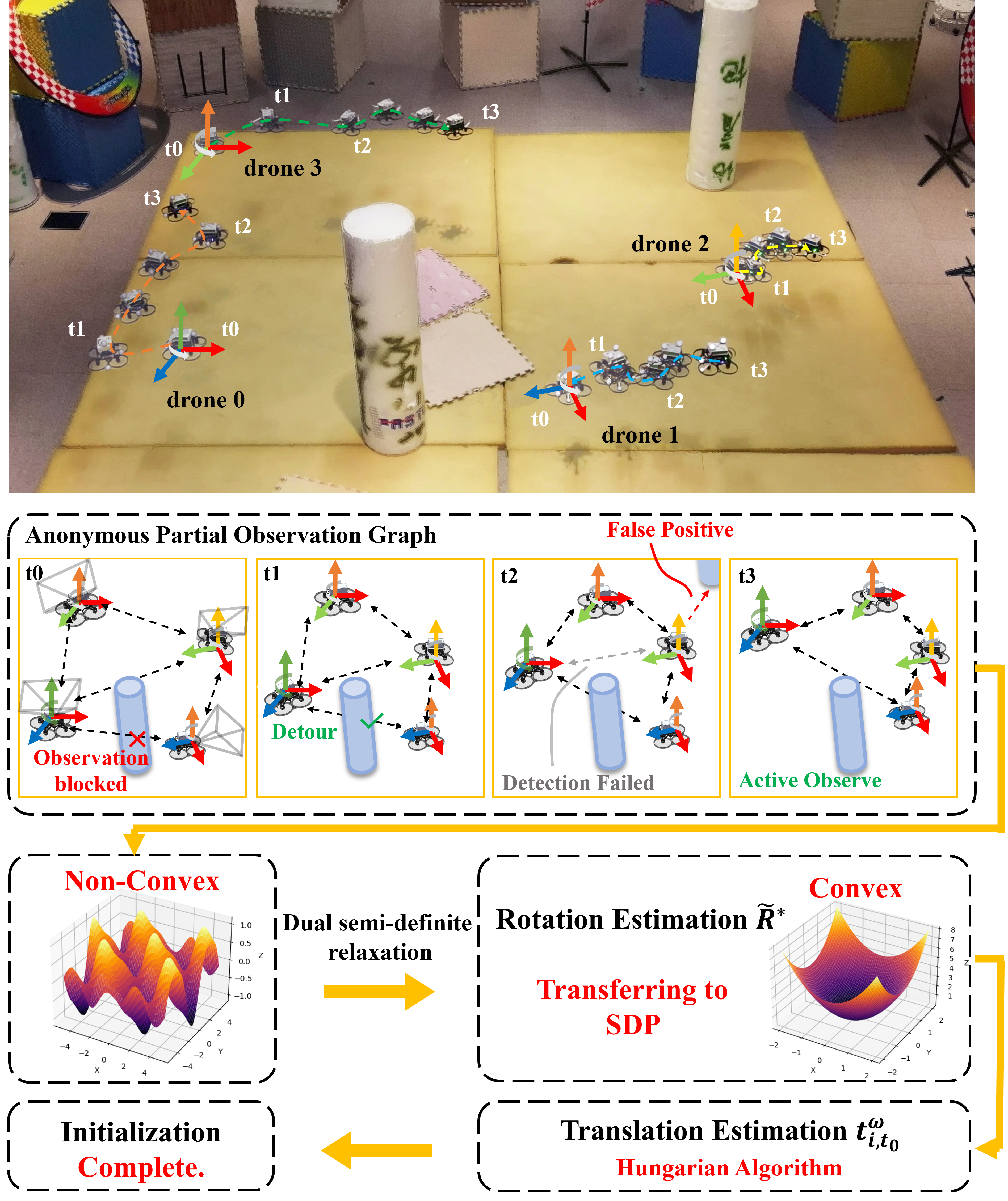}
  \caption{Schematic of vision-based coordinate initialization. Observations were sketched with bi-directional arrows. We utilize the dual semi-definite relaxation to formulate rotation estimation as a semi-definite programming (SDP) problem. Hungarian algorithm is used to solve the correspondence and come up with the translation estimations.}
  \label{schematic}
\end{figure}

Following the stereo-vision setting, there are primarily two approaches for relative pose estimation: map-based localization which involves sharing environmental features, and mutual localization, which uses inter-robot measurements. The former use common features to estimate poses, requiring extensive computation and high communication bandwidth. The latter, which do not impose significant demands on computation, communication, or textures of environments, leverage robot-to-robot range or bearing measurements to recover the initial relative pose. 

\begin{figure}[t]
  \centering
  \includegraphics[width=1\linewidth]{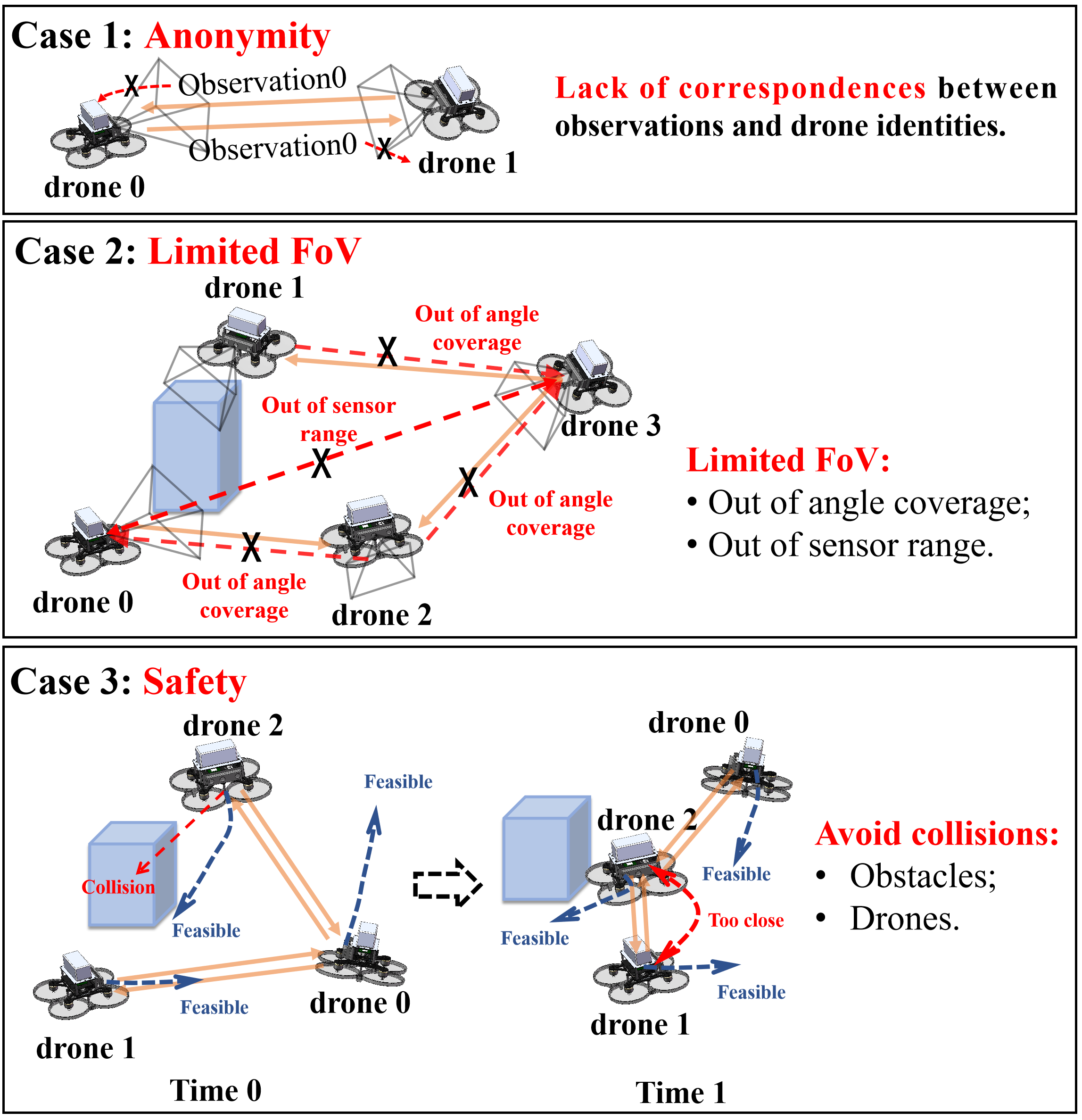}
  \caption{Challenges posed by anonymous, vision-based measurements.}
  \label{challenges}
\end{figure}

This paper primarily focuses on coordinate initialization for SWaP-constrained drone swarms with anonymous vision-based mutual observations, with no further sensor assistances like Ultra-wideband (UWB)~\cite{zhou2022swarm}, motion capture systems~\cite{trajectory-planning} or specially designed markers~\cite{wang2023bearing, wang2023Certifiably}. However, as shown in Fig.~\ref{challenges}, the vision-based measurements introduce notable challenges that are summarized as follows: 1) Limited field-of-view (FoV), which restricts the visual coverage angle and limits the sensor range, resulting in absence of mutual observations between certain robots; 2) Anonymity, which indicates a lack of correspondence between measurements and visual-inertial odometry (VIO) among other robots; And 3) safety, which requires the drones to avoid collisions with obstacles and other robots in the swarm when performing the initialization sequence.

Given the aforementioned challenges, we propose a system that can achieve coordinate initialization fast and robustly. The system consists of three parts: observation, pose estimation and active planning. Firstly, drones rotate in place while the observation module detects surrounding robots to compensate for limited FoV. Then, to recover relative rotations, fusing onboard VIO with anonymous vision-based relative measurements, we solve a semi-definite programming (SDP) problem which is relaxed from a non-convex problem. Next, we utilize the Hungarian algorithm to solve relative translations together with correspondences. Finally, the active planning module will select target positions to move to for next observation and generate feasible trajectories towards them.

Our system is validated through experiments using anonymous vision-based drone swarms in environments with obstacles, showcasing its effectiveness and robustness under different noise levels in the face of the challenges mentioned above, even with a few false positives in detection. Most importantly, compared to local optimization methods, our method can steadily obtain the global optimal solutions and reduce solving time by tenfold. Equipped with the proposed system, users can randomly place the drones anywhere without considering obstacles or adding extra sensors, and then just need to press the takeoff button. Our software handles everything needed for relative pose initialization.

Overall, our contributions are as follows:\\
\noindent
• We develop a systematic solution that integrates identification, observation, active planning, and the solving of initial relative poses between all drones with SWaP constraints;\\
\noindent
• We propose a method to first transform the mutual localization problem with bearing and range measurements into an SDP problem among a convex set that achieves global minimization;\\
\noindent
• We open source the implementation code of our method in C++ for the community\footnote{\url{https://github.com/ZJU-FAST-Lab/FACT-Coordinate-Initialization}}.

\section{RELATED WORK}
There are mainly two ways to solve the Initial mutual localization problem to get the initial relative pose between robots. They are the map-based localization methods and mutual localization methods. 

Map-based localization methods require common features to estimate relative poses. Most research~\cite{C2TAM,DOOR-SLAM} concentrates on the method. The successful implementation of these methods relies on first determining if multiple robots in the swarms have observed the same scene to acquire identical features using loop detection technique~\cite{loopdetection}. As a result, the methods require significant computation and their precision depends on the abundance of common features from the same scene between different robots, which can lead to degradation in environments with uniform or limited texture details. Furthermore, the transmission of observed data between robots, such as multiple photo frames, demands a communication network with a high bandwidth, which is not conducive to practical application.

\begin{figure*}[t]
  \centering
  \includegraphics[width=1\textwidth]{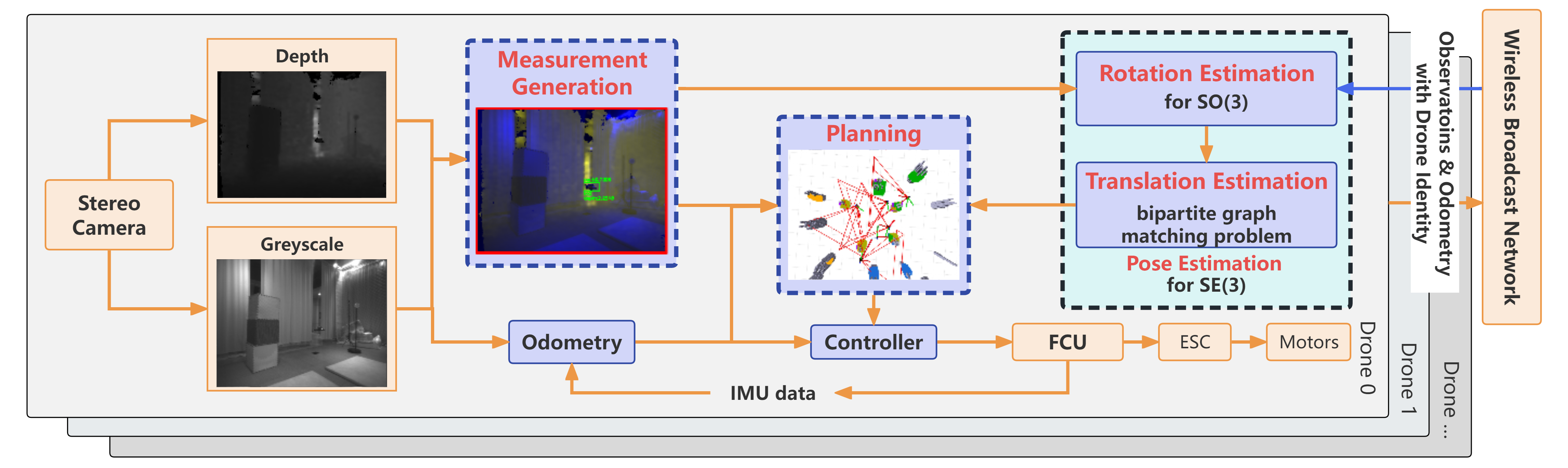}
  \caption{System overview. The main modules that our work focuses on are marked red and encapsulated by dashed lines.}
  \label{SystemOverview}
\end{figure*}

Mutual localization methods mostly need bearing or range measurements between robots. Works exclusively using range measurements, often reliant on UWB devices~\cite{range1, range2}, are constrained by the devices’ accuracy and their ability to precisely recover the relative poses of all robots. While globally optimal solutions for bearing-only measurements have been developed~\cite{wang2023bearing, wang2023Certifiably}, their reliance on bulky observation equipment renders them less viable for platforms subject to strict SWaP constraints, such as drones. Early works~\cite{bearingrange1, bearingrange2}, combining bearing and range, uses the extended Kalman filter (EKF) to estimate pose. Despite this, these studies assume a predetermined relationship between measurements and estimated poses, a premise that is frequently not applicable in real-world scenarios.

The relative pose estimation problem for swarms with anonymous mutual localization methods is first tackled by Franchi~\cite{solvability, Franchi2009MutualLI, Franchi2013MutualLI}. Franchi et al.~\cite{Franchi2009MutualLI} propose a two-phase localization framework with a multi-hypothesis extended Kalman filter (DAEKF) after the registration algorithm. The registration algorithm establishes the correspondence between observations and the identities of observed robots. Its successive work~\cite{ Franchi2013MutualLI} replaces DAEKF with particle filters (PF). Dong~\cite{Dong2015expectationmaximization} formulates a multi-robot pose graph problem and recovers initial relative poses using expectation-maximization (EM). However, DAEKF, particle filters, and EM extend computation. To better estimate relative pose, Nguyen~\cite{Nguyen2020Visionbased} improved the coupled probabilistic data association filter with the stereo camera and VIO. However, This approach struggles with the challenge of partial observations and, similar to methods employing filters, tends to have a lower success rate due to its reliance on probabilistic modeling. Compared to the above work, our method can achieve global optimization and recover relative pose at a higher success rate in the face of the five challenges mentioned above in Section~\ref{sec1}. 

\section{SYSTEM OVERVIEW}
Our work introduces a system adept at tackling the key challenges mentioned in Section~\ref{sec1}, enabling drone swarms to ascertain initial relative rotations and translations between drones in real-time, sans external localization. To promote readers' comprehension, Fig.~\ref{SystemOverview} visualizes the structure of our drone swarm system, while Algorithm~\ref{alg} logically expounds upon the comprehensive execution process of the system.

\begin{algorithm}[ht]
\caption{FACT Coordinate Initialization}
\label{alg}
\SetKwInput{KwInput}{Input}                
\SetKwInput{KwOutput}{Output}              
\SetKwFunction{FMain}{Main}                
\SetKwProg{Fn}{Function}{:}{end}           

\DontPrintSemicolon                        
\KwInput{Grayscale image, depth image, and IMU data for each drone}
\KwOutput{Initial relative positions and attitudes of all drones}

Take off to a random height.\;
\While{optimal value of Equ~\eqref{eq3} $>$ threshold}{
    Rotate in place to detect surrounding drones.\;
    Broadcast observations and VIO with identity by camera images.\;
    \If{num\_observations $\geq$ min\_num\_observations}{
            Obtain $\mathbf{R}$ by solving~\eqref{SDP} with MOSEK Fusion.\;
            Obtain $\mathbf{t}$ by solving~\eqref{eq21} with Hungarian algorithm.\;
    }
    Active planning.\;
}
Return $\mathbf{R},\mathbf{t}$
\end{algorithm}

Firstly, in the absence of external localization data, we employ an improved version of VINS-Mono~\cite{VINS-Mono} for precise localization, integrating stereo cameras and IMUs. Secondly, to precisely detect drones by depth and color images online only with onboard computing resource, we develop a lightweight deep neural network dubbed the DG-VDT module, detailed in Section~\ref{DGVDTmodule}. Moreover, to effectively solve the estimation problem mentioned in Section~\ref{problem formulation}, the solver requires quick access to valid observations. Within the limited FoV, it becomes crucial to observe as many drones as possible. Further complicating matters, drones must utilize their environmental data and observations to navigate for subsequent observation. This navigation must prioritize safety and avoid creating formation patterns with rotation symmetries. To overcome this challenge, we introduce a strategy for active and safe planning, detailed in Section~\ref{ActivePlanning}. Most importantly, we disassemble the original non-convex problem of pose estimation into two parts: orientation and translation estimation, detailed in Section~\ref{problem formulation}. We transform the original orientation estimation problem into an SDP problem through relaxation, which can solve the global optimal relative rotations, detailed in Section~\ref{rotation}. And based on the results, the translation estimation problem is solved by the Hungarian algorithm, detailed in Section~\ref{translation}. 

\section{ANONYMOUS VISION-BASED MEASUREMENT GENERATION}
\label{DGVDTmodule}
Drawing inspiration from the works of Xu et al.~\cite{xu2022omni} and Carrio et al.~\cite{carrio2018drone}, we develop a visual drone tracking module named DG-VDT that utilizes depth and gray-scale images captured by visual sensors to estimate the relative poses of drones within the swarm. DG-VDT provides detections for all drones within the FoV at each timestamp. Each detection is accompanied by a tracking ID, ensuring correspondence between consecutive frames for drone identification.

\begin{figure}[t]
  \centering
  \includegraphics[width=1\linewidth]{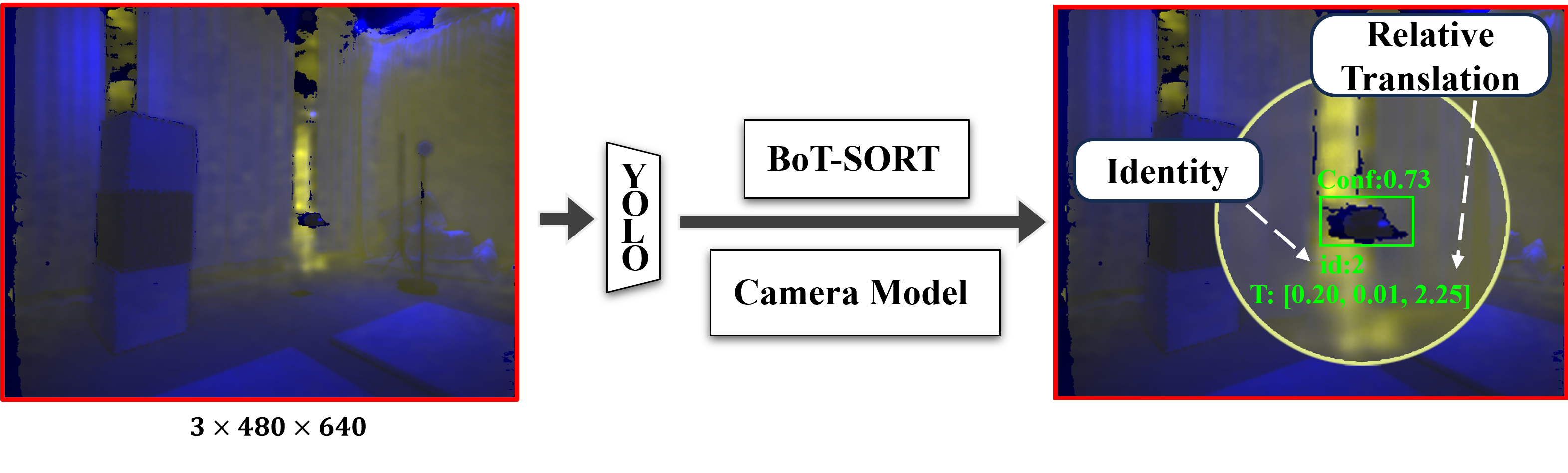}
  \caption{The implementation of the DG-VDT module.}
  \label{dg-vdt}
\end{figure}

For detection of drones, we train a customized neural network with the YOLOv8~\cite{Jocher_Ultralytics_YOLO_2023} framework, as Fig.~\ref{dg-vdt} shows. The training dataset, collected under real-world conditions, undergoes an initial automatic annotation process using the NOKOV\footnote{https://www.nokov.com} motion capture system, followed by few manual refinements. The input to the neural network consists of duplicate channels of the gray-scale image combined with the depth image, resulting in a three-channel representation. 

After obtaining the 2D bounding boxes with the network, we employ the efficient and robust tracking approach BoT-SORT~\cite{aharon2022bot} to track the visual drone targets and assign tracking IDs in DG-VDT.

To recover relative depth, we utilize a simple yet effective method of selecting a percentage threshold as effective depth information~\cite{carrio2018drone}. The method calculates the mean of the selected percentage of smallest depths within the 2D bounding box as the estimation result.

In practice, using the camera model, we reproject the center of the 2D bounding box to a 3D vector, which serves as the anonymous observation. To overcome limited FoVs, the drones rotate in place while conducting observations of surrounding robots.

\section{MUTUAL LOCALIZATION}
\subsection{Problem Formulation}
\label{problem formulation}
In this section, we formulate the problem of relative pose estimation. For the reader’s convenience, the main symbols are collected in Table~\ref{notes}.
\begin{table}[htbp]
    \renewcommand\arraystretch{1.5}
    \centering
    \caption{Main symbols used in the paper}
    \label{notes}
    \begin{tabular}{c c}
        \hline
        $d_i$ & $i$th drone \\
        $N$ & total number of drones \\
        $N_O$ & total number of observations \\
        $O_i$ & number of observed drones by $d_i$ \\
        $O_{j,t_k}$ & number of observed drones by $d_j$ at timestamp $t_k$ \\
        $f_i$ & fixed frame of drone $d_i$, defined at $t_0$ \\
        $f_w$ & fixed world frame \\
        $\mathbf{R}_{i,t_0}^w$ & relative rotation of $d_i$ w.r.t $f_w$ at timestamp $t_0$ \\
        $\mathbf{R}_{i,t_k}^{t_0}$ & relative rotation of $d_i$ w.r.t $f_i$ at timestamp $t_k$ \\
        $\mathbf{t}_{i,t_0}^w$ & relative translation of $f_i$ w.r.t $f_w$ at timestamp $t_0$ \\
        $\Vec{p}_{t_k,i}^j$ & $i$th observation by $d_j$ at timestamp $t_k$ \\
        \hline
    \end{tabular}
\end{table}

Suppose we have a team of N-drones denoted as $\{d_0, d_1,\cdots\\, d_{N-1}\}$. Each robot is attached to a fixed frame $f_i$ where the $Z_i$ axis of the frame is on the opposite direction of gravity. For convenience, we define the world coordinate $f_w$ to be the same as $f_0$. The problem of mutual localization can be defined as finding the rotation $\mathbf{R}_{i,t_0}^w \in SO(3)$ and translation $\mathbf{t}_{i,t_0}^w \in \mathbb{R}^3$ between all $f_i$ and the world coordinate $f_w$ at the initial timestamp $t_0$.

For $i,j \in \{0,\cdots N-1\},i \neq j$, if $d_i$ is detected by $d_j$ at timestamp $t_k$, we denote the vector pointing from the origin of $f_j$ to that of $f_i$ under $f_j$ as an observation $\Vec{p}_{t_k,i}^j$. Let $N_O$ denotes the total number of observations. We assume that all observations are mutual, i.e. $\forall \Vec{p}_{t_k,i}^j, \exists \Vec{p}_{t_k,j}^i$ where $k \in \{0,\cdots N_O-1\}$.

\begin{figure}[h]
  \centering
  \includegraphics[width=1\linewidth]{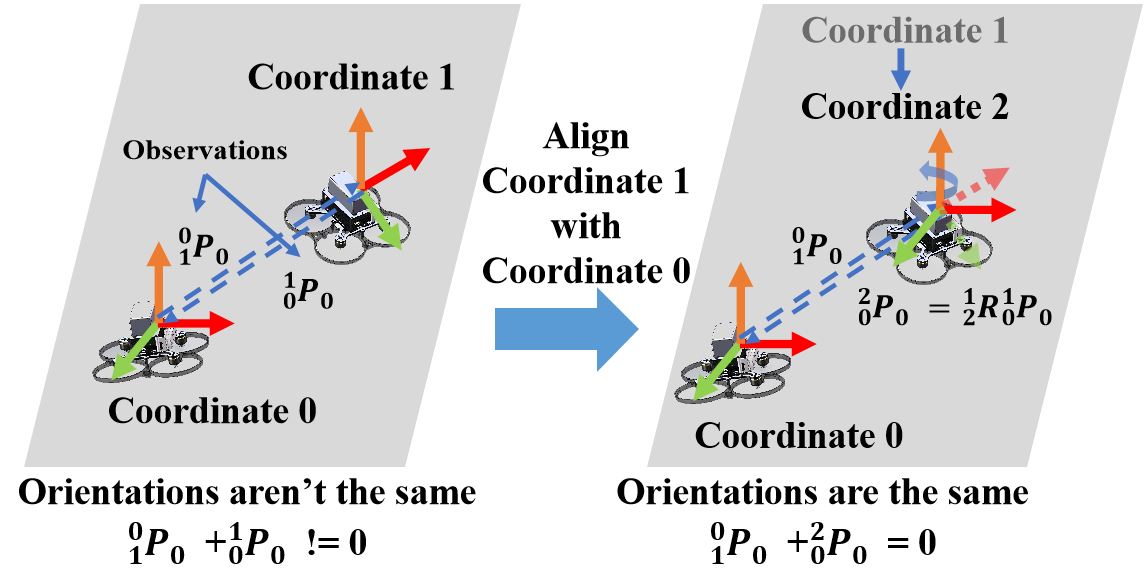}
  \caption{Schematic for a pair of mutual observations.}
  \label{mutual observations}
\end{figure}

For the problem of relative rotation estimation, let's first consider a pair of mutual observations $\Vec{p}_{t_k,i}^j, \Vec{p}_{t_k,j}^i$ as shown in Fig.~\ref{mutual observations}, which should meet

\begin{equation}
\label{eq1}
    (\mathbf{R}_{j,t_0}^w\mathbf{R}_{j,t_k}^{t_0})^T \Vec{p}_{t_k,i}^j = - (\mathbf{R}_{i,t_0}^w\mathbf{R}_{i,t_k}^{t_0}) ^T \Vec{p}_{t_k,j}^i,
\end{equation}
where $\mathbf{R}_{i,t_0}^w$ denotes the initial rotation matrix of $d_i$ relative to the world frame $f_w$, and $\mathbf{R}_{i,t_k}^{t_0}$ denotes the rotation matrix of $d_i$ at timestamp $k$ relative to $f_{i,t_0}$.

Since only the rotation along $Z$ axis is unobservable in a VIO system, we only considers the rotation along the $Z$ axes of drones, assuming the rotation along $X$ and $Y$ axes, i.e. the roll and pitch motion of drones are known. Moreover, we also assume that all $f_i$ were only different in the rotation along $Z$ axes. With the simplification, we can reduce the dimension of the rotation matrix to $\mathbf{R}_{i,t_0}^w, \mathbf{R}_{i,t_k}^{t_0} \in SO(2)$. If we consider all mutual observations at a given timestamp $t_k$, we can derive the error of rotation estimation as

\begin{equation}
    e_{t_k} = \sum_{j=0}^{N-1} (\mathbf{R}_{t_0} \mathbf{C}_j \mathbf{R}_{t_k}^{t_0}\mathbf{C}_j)^T\sum_{i\in O_{j,t_k}} \Vec{p}_{t_k,i}^j,
\end{equation}
where, similar to~\cite{wang2023bearing}, $\mathbf{R}_{t_0} := [\mathbf{R}_{0,t_0}^w,\mathbf{R}_{1,t_0}^w,\cdots \mathbf{R}_{N-1,t_0}^w]\in SO(2)^N$ is our decision variable, $\mathbf{C}_j = e_j \otimes I_3$ is the choose matrix for robot $d_j$, $\mathbf{R}_{t_k}^{t_0} := [\mathbf{R}_{0,t_k}^{t_0},\mathbf{R}_{1,t_k}^{t_0},\cdots,\mathbf{R}_{N-1,t_k}^{t_0}]$ can be acquired through the odometry of robots and $O_{j,t_k}$ is the observed drones by $d_j$ at timestamp $t_k$. Note that items in $O_{j,t_k}$ are abstract numbers rather than specific identities of drones, as the visual detection we use is anonymous.

Finally, considering all the timestamps, we can formulate the problem of relative rotation estimation as

\begin{equation}
\label{eq3}
    \mathbf{R}_{t_0}^* = \mathop{\arg\min}\limits_{\mathbf{R}_{t_0}\in SO(2)^N} \sum_{k=0}^{N_O-1} e_{t_k}^Te_{t_k}.
\end{equation}

As for the problem relative translation estimation, recall that in the above derivations, the identities of the observed drone $i$ in $\Vec{p}_{t_k,i}^j$ are unknown. Therefore, to estimate the translations $\mathbf{t}_{i,t_0}^w$, we have to find the correspondence between the observations and the identities of drones. The problem can be defined as identifying a binary matrix $\mathbf{A}_i \in \mathbb{R}^{O_i\times N}$
for each drone $d_i$, where $O_i$ denotes the total number of observed planes by $d_i$, each element $^ia_x^y$ is a binary value, $^ia_x^y = 1$ if and only if the $x$th observation of $d_i$ is $d_y$. As every observation corresponds to a certain drone, $\sum_x {^ia_x^y} = 1, \forall y\in \{0,\cdots,N-1\}$. Considering partial observations, we have $\sum_y {^ia_x^y}\in \{0,1\}, \forall x\in \{0,\cdots,O_i-1\}$, where $\sum_y {^ia_x^y}=0$ indicates that $d_y$ is not observed by $d_i$. With the matrix, we define a correspondence function

\begin{equation}
    \mathcal{C}(\alpha, i) = \beta \mid {^ia_\beta^\alpha} = 1,
\end{equation}
which find the corresponding number $\beta$ for drone $d_\alpha$ in $O_i$. Then for drones directly observed by $d_0$, we have

\begin{equation}
\label{eq6}
    \mathbf{t}_{i,t_0}^w = {\mathbf{R}_{1, t_k}^{t_0}}^T  \Vec{p}_{t_k,\mathcal{C}(i,1)}^1.
\end{equation}

As for the drones not directly observed by $d_0$, we can use the following equation:

\begin{equation}
\label{eq7}
    \Vec{p}_{t_0, \beta}^\alpha = \Vec{p}_{t_0, \gamma}^\alpha + (\mathbf{R}_{\gamma,t_0}^{w} \mathbf{R}_{\gamma,t_k}^{t_0})^T \Vec{p}_{t_k,\mathcal{C}(\beta, \gamma)}^\gamma,
\end{equation}
where $d_\gamma$ is observed by $d_\alpha$ to find $\Vec{p}_{t_0, \beta}^\alpha$ for $d_\beta$ that is not directly observed by $d_\alpha$. With the equation, we can iteratively calculate $\Vec{p}_{t_k,\mathcal{C}(i,1)}^1$ for any $i\in N$ to obtain all $\mathbf{t}_{i,t_0}^w$.

\subsection{Rotation Estimation}
\label{rotation}
If we use $\mathbf{R}(\theta)$ to denote the corresponding rotation matrix of turning $\theta$ along the $Z$ axis, the properties of rotation matrix indicates that $\mathbf{R}(\theta)^T = \mathbf{R}(-\theta)$. Denoting $\mathbf{R}_{t_0}$ and $\mathbf{R}_{t_k}^{t_0}$ with $\mathbf{R}(\Theta_0)$ and $\mathbf{R}(\Theta_k^0)$, we can derive the target function in~\eqref{eq3} as

\begin{align}
    &\sum_{k=0}^{N_O-1} e_{t_k}^Te_{t_k} \\
    =&\sum_{k=0}^{N_O-1}  \sum_{m=0}^{N-1} \sum_{n=0}^{N-1} (\mathbf{R}(-\Theta_0) \mathbf{C}_m \mathcal{P}_k^m)^T (\mathbf{R}(-\Theta_0) \mathbf{C}_n \mathcal{P}_k^n) \\
    =&tr\big(\sum_{k=0}^{N_O-1} \sum_{m=0}^{N-1} \sum_{n=0}^{N-1}  \mathbf{C}_n \mathcal{P}_k^n {\mathcal{P}_k^m}^T \mathbf{C}_m^T \mathbf{R}(-\Theta_0)^T \mathbf{R}(-\Theta_0) \big),
\end{align}
where

\begin{equation}
    \mathcal{P}_k^j = \mathbf{R}(-\Theta_k^0) \mathbf{C}_j \sum_{i\in O_{j,t_k}}\Vec{p}_{t_k,i}^j
\end{equation}
is a constant since $\mathbf{C}_j$ is known and $\mathbf{R}(-\Theta_k^0)$ can be acquired using onboard odometry. Let 

\begin{equation}
    \mathcal{Q}(k,N,N_O) = \sum_{k=0}^{N_O-1} \sum_{m=0}^{N-1} \sum_{n=0}^{N-1}  \mathbf{C}_n \mathcal{P}_k^n {\mathcal{P}_k^m}^T \mathbf{C}_m^T,
\end{equation}
which is a constant, we can redefine the problem of rotation estimation as

\begin{equation}
\label{rotation estimation}
    \mathbf{R}_{t_0}^* = \mathop{\arg\min}\limits_{\mathbf{R}_{t_0}\in SO(2)^N} tr\big(\mathcal{Q}(k,N,N_O) \mathbf{R}(-\Theta_0)^T \mathbf{R}(-\Theta_0)\big).
\end{equation}

Now let
\begin{equation}
    \mathbf{Z} = \mathbf{R}(-\Theta_0)^T \mathbf{R}(-\Theta_0)
\end{equation}
be the new decision variable, according to SE-Sync~\cite{rosen2019se}, utilizing the dual semi-definite relaxation, we can redefine the problem of rotation estimation as

\begin{align}
\label{SDP}
    \mathbf{Z}^* &= \mathop{\arg\min}\limits_{\mathbf{Z}\in Sym(2N)} tr(\mathcal{Q}_{k,N,N_O} \mathbf{Z}) \\
    s.t.\hspace{3mm} &\mathbf{Z} = \left[
    \begin{array}{cccc}
        I_2 & * & \cdots & * \\
        * & I_2 & \cdots & * \\
        \hdots & \hdots & \ddots & \hdots \\
        * & * & \cdots & I_2\\
    \end{array}
    \right] \succeq 0,
\end{align}
where $\mathbf{Z} \succeq 0$ means that $\mathbf{Z}$ is a positive semi-definite matrix, indicating that the problem is a standard SDP problem that could be solved with numerical solvers. 

After obtaining $\mathbf{Z}^*$, we can calculate $\mathbf{R}_{t_0}^*$ with

\begin{equation}
    \mathbf{R}_{t_0}^* = \mathbf{U} * \sqrt{\mathbf{S}},
\end{equation}
where $\mathbf{U}$ and $\mathbf{S}$ are the results of singular value decomposition (SVD) of $\mathbf{Z}^*$. As our previous work~\cite{wang2023Certifiably} indicates, if $rank(\mathbf{Z}^*) \le N+1$, complete observations is acquired and global optimality of $\mathbf{R}_{t_0}^*$ can be achieved.

\subsection{Translation Estimation}
\label{translation}
We've already defined the problem of mutual translation estimation as finding a correspondence matrix $\mathbf{A}_i$ for all drones in Section~\ref{problem formulation}. To solve the problem, let's first consider the matching error between two observations $\Vec{p}_{t_k,j}^i$ of drone $d_i$ and $\Vec{p}_{t_k,y}^x$ of drone $d_x$ at timestamp $t_k$:

\begin{equation}
    ^{i,j}e_{xy} = \big| [\mathbf{R}_{i,t_0}^w\mathbf{R}_{i,t_k}^{t_0}]^T \Vec{p}_{t_k,j}^i +[\mathbf{R}_{x,t_0}^w\mathbf{R}_{x,t_k}^{t_0}] ^T \Vec{p}_{t_k,y}^x\big|.
\end{equation}

As Equation~\eqref{eq1} shows, if the observations were a set of mutual observations, i.e. the $j$th observation of $d_i$ is $d_x$ and the $y$th observation of $d_x$ is $d_i$, the error term should be zero. Utilizing the matching error, we define a matching error matrix for a drone $d_i$ as

\begin{equation}
    \mathbf{E_i} = \left[
    \begin{array}{cccc}
        ^{i,1}\mathbf{e}_1 & ^{i,1}\mathbf{e}_2 & \cdots & ^{i,1}\mathbf{e}_N \\
        ^{i,2}\mathbf{e}_1 & ^{i,2}\mathbf{e}_2 & \cdots & ^{i,2}\mathbf{e}_N \\
        \vdots & \vdots & \ddots & \vdots \\
        ^{i,O_i}\mathbf{e}_1 & ^{i,O_i}\mathbf{e}_2 & \cdots & ^{i,O_i}\mathbf{e}_N \\
    \end{array}
    \right]
    \in \mathbb{R}^{Obs_i \times O_i},
\end{equation}
where $^{i,j}\mathbf{e}_k = [^{i,j}e_{k1}, ^{i,j}e_{k2},\cdots,^{i,j}e_{kO_k}]$ is the array of matching errors between all observations of drone $d_k$ and the $j$th observation of $d_i$, $Obs_i$ represents the total number of observations except those of $d_i$. Using $\odot$ to represent an element-wise multiplication of two matrixes, we can obtain $\mathbf{A}_i$ as the result of

\begin{align}
\label{eq21}
    \mathbf{A}_i^* &= \mathop{\arg\min}_{\mathbf{A}_i} \mathbf{A}_i \odot \mathbf{E}_i \\
    s.t.\hspace{3mm}&\sum_x {^ia_x^y} = 1, \forall y\in \{0,\cdots,N-1\}, \\
    &\sum_y {^ia_x^y}\in\{0,1\}, \forall x\in \{0,\cdots,O_i-1\}, \\
    &a_x^y \in \{0,1\},
\end{align}
which is a bipartite graph matching problem that could be solved using the Hungarian algorithm. Utilizing the algorithm to all drones, we can obtain $\mathbf{A}_i$ for every $d_i$, resulting in a directed graph containing all drones and their observations. Then, we apply depth first search in the graph to obtain paths from $d_1$ to all other drones and proceed to use Equation~\eqref{eq6} and~\eqref{eq7} to estimate translation $\mathbf{t}_{i,t_0}^w$ for all $i\in \{0,\cdots,N-1\}$.

\section{ACTIVE AND SAFE PLANNING}
\label{ActivePlanning}
To estimate the relative pose precisely, our algorithm needs multiple observations from distinct positions. Therefore, we introduce a strategy aiming at finding ideal positions for potential observations quickly and generating secure trajectories towards the selected targets with only local environment data and observations. The overall process is as follows.

Firstly, drones tend to veer off course during rotation due to VIO drifts and limited control accuracy. Therefore, to prevent collisions caused by unpredictable position drifts, we set a fixed minimum distance between drones for safety. Secondly, since the relative poses between drones are still unknown during initialization, drones can only plan with locally obtained information. Thus, to avoid conflicts in planned trajectories, we set a maximum moving distance based on the position of the closest drone. Thirdly, the module assesses the distribution of surrounding drones and randomly selects between exploring towards a random direction with a small likelihood and moving towards the farthest drone. Next, our algorithm will verify if the intended destination falls within an obstacle. If so, the target point will be adjusted to the nearest valid position outside of the obstacle. Lastly, we employ a planner to plan trajectories to reach the targets for the next observation.

\section{BENCHMARK AND EXPERIMENT}
In this section, we first compare the rotation estimation of our algorithm with local optimization methods in simulated environments with different numbers of drones under various levels of noise. Since the precision of translation estimation mainly depends on that of rotation estimation, we will focus on the results of rotation estimation. Furthermore, to validate the practicality and robustness in the real world, we employ the Nvidia Orin NX for the onboard computation, and the Intel Realsense 430 camera for the vision sensing, deploying our algorithm on a vision-based drone swarm whose one single drone weighs just 450 grams. To generate trajectories, we deploy EGO-Planner~\cite{ego-swarm} to avoid collisions. Our algorithm is implemented in C++ using the MOSEK Fusion API.

\subsection{Simulated Comparison with Local Optimization Methods}
\begin{figure}[b]
    \centering
    \includegraphics[width=1\linewidth]{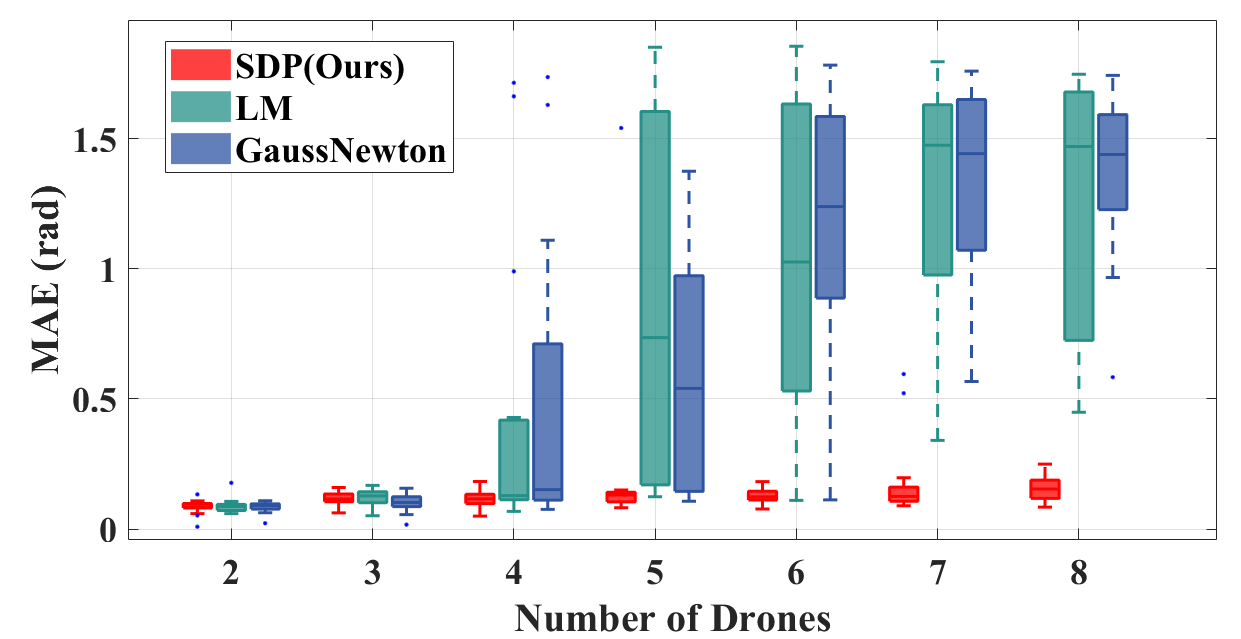}
    \caption{Sandbox figures of mean absolute error. Each method is tested with 2 to 8 drones respectively for 20 times.}
    \label{sim result}
\end{figure}
\begin{table}[b]
    \centering
    \caption{Time consumed(s) by rotation estimation}
    \label{rotation solving time}
    \begin{tabular}[width=1\linewidth]{|c|c|c|c|c|c|c|c|}
        \hline
        \textbf{Num} & \textbf{2} & \textbf{3} & \textbf{4} & \textbf{5} & \textbf{6} & \textbf{7} & \textbf{8} \\
        \hline
        \textbf{SDP} & \textbf{0.004} & \textbf{0.009} & \textbf{0.021} & \textbf{0.032} & \textbf{0.109} & \textbf{0.163} & \textbf{0.253} \\
        \hline
        LM & 0.005 & 0.035 & 0.084 & 0.264 & 0.408 & 0.854 & 2.043 \\
        \hline
        GN & 0.012 & 0.034 & 0.081 & 0.271 & 0.388 & 1.016 & 3.434 \\
        \hline
    \end{tabular}
\end{table}
\begin{figure*}[t]
    \centering
    \subfigure[Real-world experiments with 4 drones.]{
        \begin{minipage}{0.49\textwidth}
            \centering
            \includegraphics[width=1\textwidth]{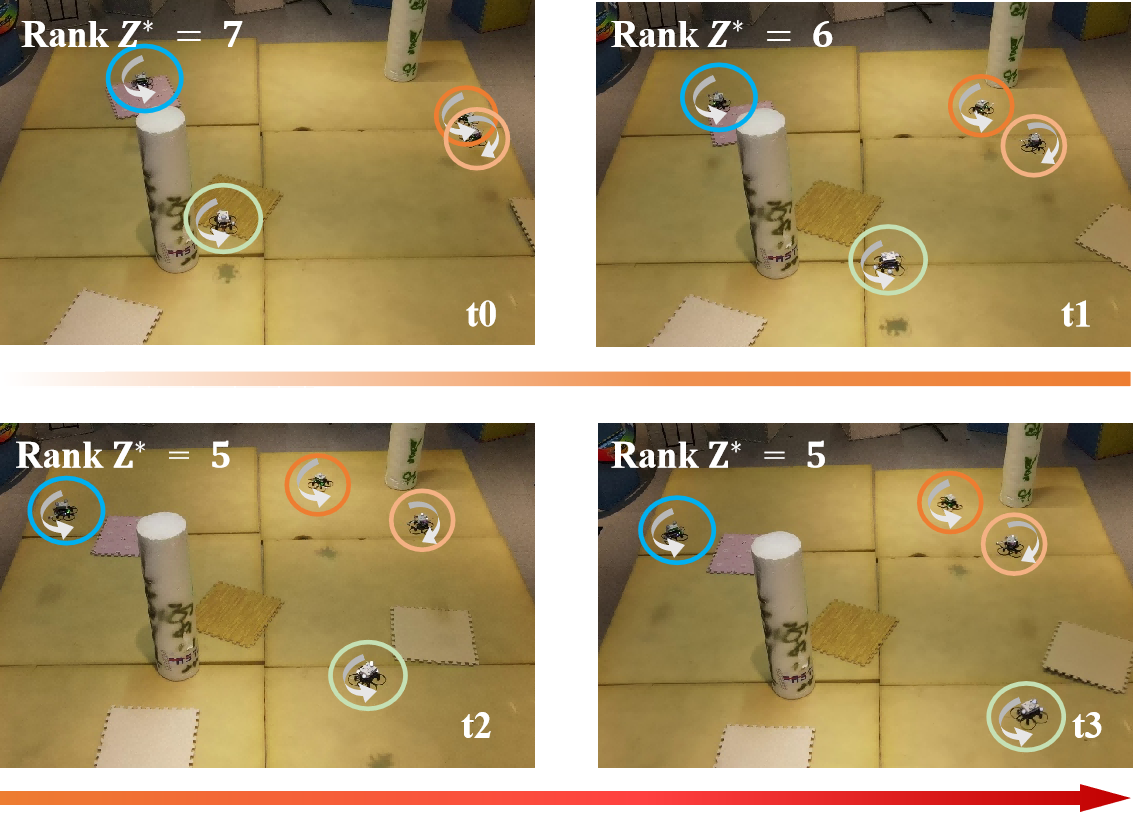}
            \label{4 drone experiment}
        \end{minipage}
    }
    \subfigure[Real-world experiments with 3 drones.]{
        \begin{minipage}{0.46\textwidth}
            \centering
            \includegraphics[width=1\textwidth]{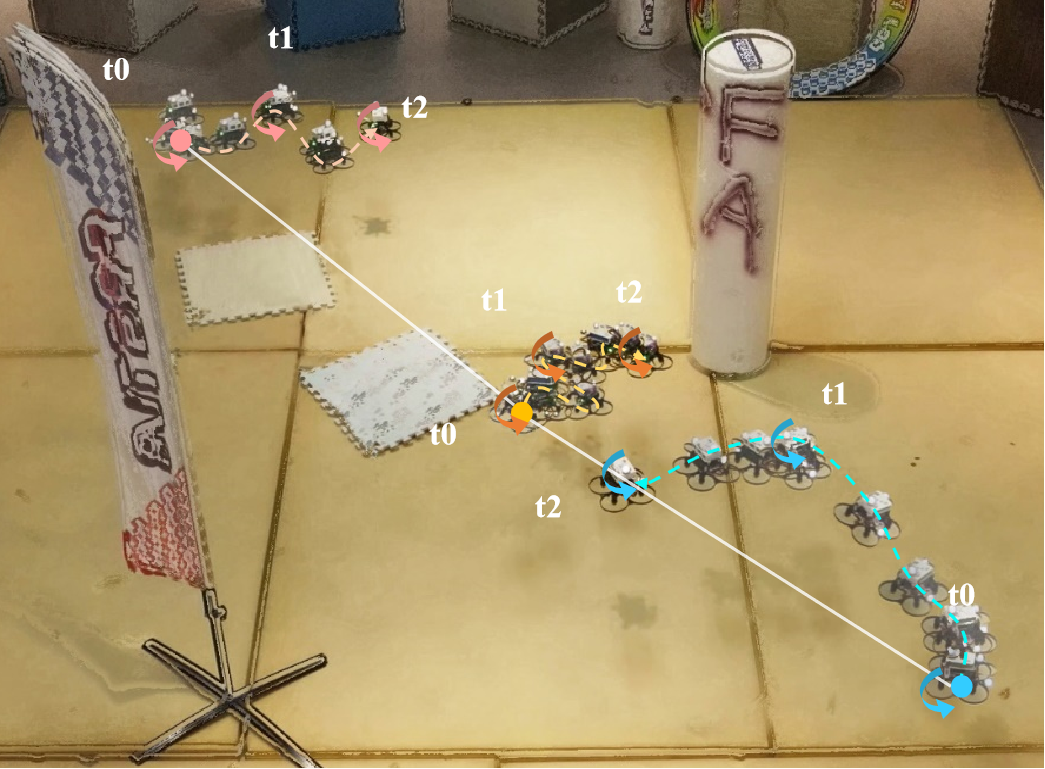}
            \label{rotation symmetry}
        \end{minipage}
    }
    \caption{Real-world experiments with different number of drones. Fig. (b) displays a case where we encounter a rotation symmetry formation.}
    \label{real experiments}
\end{figure*}

\begin{figure}[t]
    \centering
    \includegraphics[width=1\linewidth]{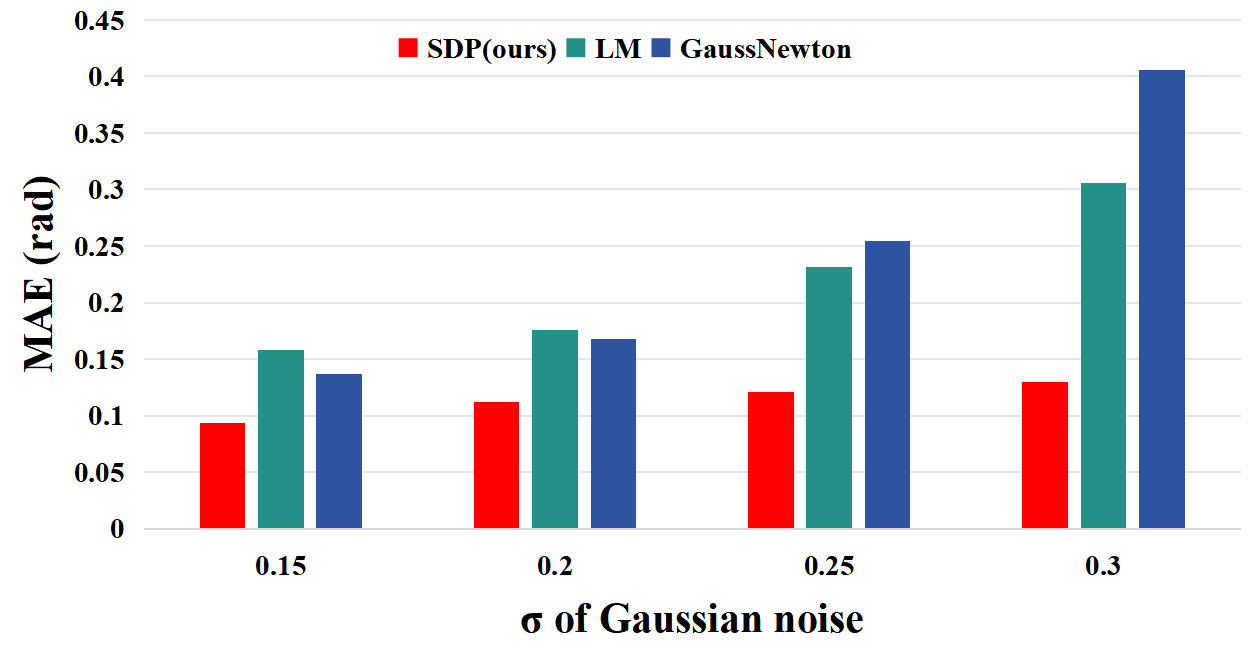}
    \caption{Mean absolute error of rotation estimation in 4-drone experiments with different noise added.}
    \label{sim result2}
\end{figure}

We first conduct experiments in simulation with 2 to 8 drones initializing randomly in the environment. To demonstrate the robustness of our system, we add various levels of Gaussian noise to the observations. For comparison, instead of transforming the rotation estimation problem~\eqref{rotation estimation} to an SDP problem~\eqref{SDP}, we utilize local optimization methods (Levenberg-Marquardt and Gauss-Newton) to solve~\eqref{rotation estimation} and conduct experiments under the same circumstances. We measure the time consumed by our method and two local optimization methods for the rotation estimation, as detailed in Table~\ref{rotation solving time}, where 'GN' denotes the Gauss-Newton method. The mean absolute error (MAE) of rotation estimation with different number of drones are shown in Fig.~\ref{sim result}. We also test the algorithms' capabilities under different level of noise, as sketched in Fig.~\ref{sim result2}.

Generally, our method can obtain global optimal values fast and resiliently. As the plot indicates, our algorithm performs well under different levels of noises, demonstrating the capability to come up with decent results under noisy observations. With 2 or 3 drones, as observations are mostly rigid or even complete~\cite{Franchi2013MutualLI}, the errors are relatively small for both our methods and local optimization ones. As the number of drones increases, partial observations and more variables start to make it harder for local optimization methods, whose errors increase rapidly as shown in Fig.~\ref{sim result}. Yet for our method, the errors gradually decrease as the number of drones increases, as the algorithm can better tolerate random disturbances with more observations while trying to find the global optimal values. Another advantage of our algorithm is a better tolerance to noises with the ability to find global minimums, while local optimization methods often get stuck in local minimums caused by noises, resulting in worse performance with higher noises as Fig.~\ref{sim result2} indicates. It's also worth mentioning that our methods' results are relatively stable, yet the local optimization methods' errors varies significantly, as shown in Fig.~\ref{sim result}.  What's more, our methods also possesses significant advantage on calculating time, especially with more number of drones, as the average solving time of our algorithm for 8 drones is $253ms$, while Levenberg-Marquardt and Gauss-Newton methods need $2.04s$ and $3.43s$ respectively.

\subsection{Real-world Experiments}
\begin{figure}[t]
    \centering
    \includegraphics[width=1\linewidth]{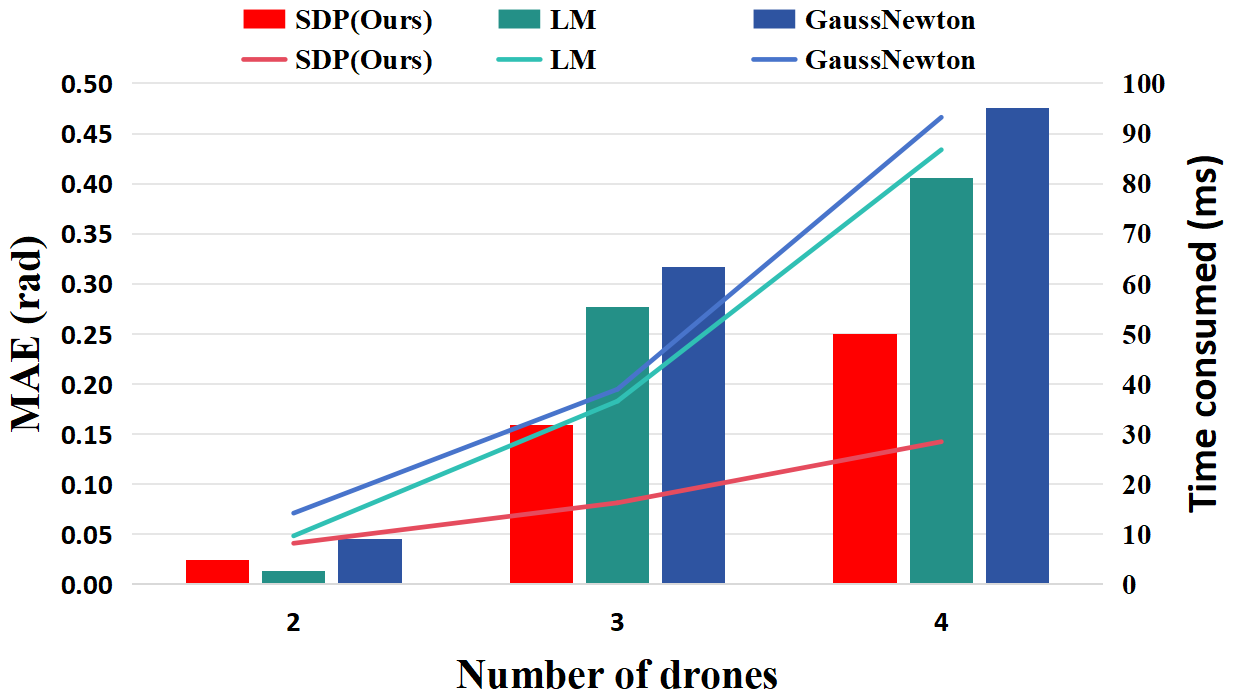}
    \caption{Mean absolute error and consumed time of rotation estimation with different methods. Errors and times are sketched with histograms and lines respectively. Ground truths are obtained by NOKOV motion capture system.}
    \label{real result}
\end{figure}

For real-world experiments, we place 2-4 drones with random poses in environments with randomly placed obstacles. For practical deployment in more real-world scenarios, we utilize VINS-Mono~\cite{VINS-Mono} for localization and the DG-VDT for detection. Similar to what we've done in simulation experiments, we compare our approach to solve the rotation estimation with local optimization methods.

In general, the system can achieve coordinate initialization fast and robustly with only onboard vision sensors under limited onboard computing resources. As Fig.~\ref{real result} indicates, our algorithm exhibits significant advantage in terms of both accuracy and time consumption with more than 2 drones, demonstrating the systems superior capability to obtain decent results under noises introduced by vision-based detections and odometries. The time consumed by the Hungarian Algorithm is around $0.4ms$, recovering relative translations in real-time.

As Fig.~\ref{4 drone experiment} shows, the rank of $Z^*$ descends from $8$ to $5$ as the number of observations increase from $0$ to $3$. The fourth observation is conducted to provide extra information, increasing the accuracy of estimations against the noises. What's more, as Fig.~\ref{rotation symmetry} shows, our system can overcome the extra challenge of rotation symmetric formations. Rotation symmetries will lead to multiple solutions with anonymous observations~\cite{solvability}. The active planning module can break rotation symmetries with designed random movements, conducting observations with different formations and thus achieving relative pose estimation even under an initially rotation symmetric formation.

\section{CONCLUSION AND FUTURE WORK}
In this paper, we propose a system which can recover all relative poses fast and robustly. It involves a global optimal algorithm that utilizes anonymous vision-based measurements to formulate an SDP problem with the dual semi-definite relaxation. For translation estimation, we use the Hungarian algorithm is utilized to solve the bipartite graph matching problem, obtaining correspondences simultaneously. To obtain complete observations, we introduce an active planning strategy to conduct multiple measurements. Extensive experiments in simulations under various noise levels together with real-world experiments demonstrate our approach's superior capacity to effectively address the challenges mentioned in Section~\ref{sec1} compared with local optimization methods. Our algorithm can steadily obtain accurate relative poses, running in real-time with onboard computers of SWaP-constrained drones. Our future work will explore and improve the threshold for observation noise to tackle uncertainties in vision-based identifications more robustly.

\bibliographystyle{ieeetr}
\bibliography{references}

\end{document}